\documentclass[11pt]{article}

\usepackage[final]{acl} 
\usepackage{times}
\usepackage{latexsym}

\usepackage[T1]{fontenc}
\usepackage[utf8]{inputenc}

\usepackage{microtype}

\usepackage{inconsolata}

\usepackage{graphicx}

\usepackage{booktabs}
\usepackage{multirow}
\usepackage{makecell}
\usepackage{tcolorbox}

\usepackage[table]{xcolor}
\usepackage{xcolor}

\usepackage{listings}

\usepackage{kotex}     % 한글 처리
\usepackage[dvipsnames]{xcolor}

\title{An Empirical Study of On-Device Translation for Real-Time Live-Stream Chat on Mobile Devices}

\author{
 \textbf{Jeiyoon Park\textsuperscript{1,}\textsuperscript{\textbf{\textdagger}}},
 \textbf{Daehwan Lee\textsuperscript{1}},
 \textbf{Changmin Yeo\textsuperscript{1}},
 \textbf{Yongshin Han\textsuperscript{1}},
 \textbf{Minseop Kim\textsuperscript{1}},
\\
 \textsuperscript{1}SOOP
\\
 k4ke@korea.ac.kr, \{ldh, yeolife, winterfeb, usio\}@sooplive.com
\\
 \small{
   \url{https://huggingface.co/datasets/naruto-soop/LiveChatBench}
 }
}

\begin{document} 
\maketitle

% 첫 페이지 하단에 † 각주 추가
\begingroup
\renewcommand{\thefootnote}{\textbf{\textdagger}}
\footnotetext[1]{%
  The first author is currently at Smilegate.
}
\endgroup

\begin{abstract}
Despite its efficiency, there has been little research on the practical aspects required for real-world deployment of on-device AI models, such as the device’s CPU utilization and thermal conditions. In this paper, through extensive experiments, we investigate two key issues that must be addressed to deploy on-device models in real-world services: (i) the selection of on-device models and the resource consumption of each model, and (ii) the capability and potential of on-device models for domain adaptation. To this end, we focus on a task of translating live-stream chat messages and manually construct \textsc{LiveChatBench}, a benchmark consisting of 1,000 Korean–English parallel sentence pairs. Experiments on five mobile devices provide a systematic empirical assessment of widely adopted on-device models, highlighting the importance of model selection and deployment constraints when adapting them to specialized tasks and serving a large and heterogeneous user base. We expect that our findings will offer practical insights into both the capabilities and limitations of on-device models for real-world AI service deployment.

\end{abstract}

\section{Introduction}
\label{sec:intro}

Large language models (LLMs) have driven major advances across a wide range of AI applications, while small language models (SLMs) have also shown significant promise by achieving competitive performance with substantially lower computational and memory demands \cite{xu2024ondevicelanguagemodelscomprehensive,bohdal-etal-2025-device,liu2024mobilellmoptimizingsubbillionparameter,allal2025smollm2smolgoesbig,pham-etal-2025-slimlm,hau2025llmsmobileappspractices,zhao2025mobilellmr1exploringlimitssubbillion}. However, although real-world deployment of on-device models requires accounting for heterogeneous device conditions and user environments, existing studies have paid limited attention to practical operational constraints, such as CPU resource consumption and fluctuations in device temperature. Though \citep{liu2024mobilellmoptimizingsubbillionparameter} and \citep{pham-etal-2025-slimlm} address practical challenges, they focus only on model architecture or training.

In this paper, we examine the challenges that must be addressed to deploy on-device models in real-world services, focusing on selecting appropriate models, characterizing the resource usage of each, and assessing their capacity and potential for domain adaptation. Inspired by a culturally-aware evaluation
benchmark \cite{kim-etal-2024-click}, we manually construct LiveChatBench, a high-quality benchmark consisting of 1,000 Korean–English live-streaming chat translation pairs that include memes, slang expressions, and ungrammatical or ill-formed text patterns, along with annotations of the knowledge required for accurate translation.

We conduct two types of experiments using three SLMs  with 270M, 0.6B, and 1B parameters \cite{gemmateam2025gemma3technicalreport,yang2025qwen3technicalreport}, three iOS devices, and two Android devices: (i) we measure mobile-device CPU utilization, temperature variations, time to first token, and runtime on LiveChatBench under CPU-only execution and under GPU-accelerated execution, and (ii) we evaluate the domain adaptation performance of on-device models on LiveChatBench, FLORES-200 \cite{nllbteam2022languageleftbehindscaling}, and WMT++ \cite{deutsch-etal-2025-wmt24}.

\begin{table*}[t]
\centering
\small
\resizebox{\textwidth}{!}{
\begin{tabular}{@{}lll@{}}
\toprule
\textbf{Source (ko)} & \textbf{Target (en)} & \textbf{Background Knowledge} \\ \midrule
\textbf{\textcolor{ForestGreen}{메접}} \textbf{\textcolor{ForestGreen}{빌드업}} 지린다 & This build-up to quitting MapleStory is insane. & \textbf{\textcolor{ForestGreen}{메접}}: '메접'은 온라인 게임 메이플스토리에서 접는다 (그만둔다)는 뜻으로 쓰이는 줄임말입니다. \\ 
 &  & \textbf{\textcolor{ForestGreen}{빌드업}}: '빌드 업(build up)'은 '쌓아 올리다', '점점 증가시키다'라는 뜻으로, 어떤 것을 단계적으로 만들어가는 과정을 의미합니다.  \\ \midrule
애가 \textbf{\textcolor{ForestGreen}{억타}}인데 & The kid is being forced to play StarCraft, though. & \textbf{\textcolor{ForestGreen}{억타}}: '억지로 스타크래프트를 한다'라는 의미. '억마크'처럼 '억+(게임 이름)'으로 응용된다. \\ \midrule
어제 \textbf{\textcolor{ForestGreen}{비챤}}님 롤대회가 인기많던데.. & Yesterday, VIichan’s League of Legends tournament seemed really popular. & \textbf{\textcolor{ForestGreen}{비챤}}: 인터넷 방송 플랫폼 "SOOP (숲)"의 스트리머 이름. \\
\bottomrule
\end{tabular}}
\caption{Examples from the LiveChatBench dataset.}
\label{tab:sample_livechatbench}
\end{table*}

Results demonstrate that, given the performance constraints of client devices and the need to preserve user convenience, deploying on-device models in real-world applications inevitably requires making a limited set of design trade-offs; nevertheless, through the lens of domain adaptation, the proposed approach achieves performance comparable to that of commercial models such as GPT-5.1 on the well-targeted task, highlighting both the effectiveness and the latent potential of on-device AI. We believe our findings offer insights into the unavoidable challenges faced by researchers designing models under strict user-side resource constraints, including limited battery capacity \cite{liu2024mobilellmoptimizingsubbillionparameter,6237004,10.1109/ISCA.2016.30}.

\section{Datasets}
\label{sec:datasets}

\subsection{Data Collection}
\label{sec:data_collection}

First, we collect approximately 30 million chat messages from a live-streaming platform\footnote{\url{https://www.sooplive.co.kr/}}\textsuperscript{,}\footnote{All data are collected and used in accordance with the relevant usage permissions, and no personally identifiable information is included. The dataset consists of Korean chat data and is used exclusively for academic research.}. Then, we filter out overly uninformative instances (e.g., \textit{“ㅋㅋㅋ”}, meaning \textit{“lol”}) and messages longer than 50 characters, since our analysis of the collected livestream chat data showed that 99.03\% of messages were within this length. Synthetic parallel pairs are generated using the pipeline described in Section \ref{sec:livechatbench}. We finally construct a dataset comprising approximately 1.5 million training and development examples in total. Note that this dataset contains a wide range of memes, slang expressions, and ungrammatical or ill-formed text patterns, which can substantially degrade an LLM’s ability to interpret and understand the content \cite{sun-etal-2022-semantically,sun-etal-2024-toward,wuraola-etal-2024-understanding}. 

\begin{table}[t]
\centering
\footnotesize
% \resizebox{0.4\textwidth}{!}{
\begin{tabular}{@{}llll@{}}
\toprule
 & \textbf{BM25} & \textbf{LLM} & \textbf{BM25 + LLM}\\ \midrule
\textbf{Micro Recall ($\uparrow$)} & 0.4834 & 0.7031 & \textbf{0.8107}\\ 
\bottomrule
\end{tabular}
% }
\caption{\textbf{Micro-averaged recall results.} The methods compared are BM25, LLM-based entity extraction (using GPT-5.1), and a hybrid BM25+LLM approach.}
\label{tab:eval_pipe}
\end{table}

\subsection{LiveChatBench}
\label{sec:livechatbench}
Building on previous studies showing that synthetic data can improve translation performance  \cite{kartik-etal-2024-synthetic,de-gibert-etal-2025-scaling}, we use an LLM to translate the dataset collected in Section \ref{sec:data_collection}. However, simply constructing a synthetic dataset is more likely to inject erroneous knowledge because chat data are difficult for even state-of-the-art LLMs to interpret. Since knowledge injection greatly facilitates the construction of synthetic data \cite{shen2025ragsynthsyntheticdatarobust}, we first manually build a dictionary of internet terminology and slang required for translation, consisting of 656 words, and then adopt a framework that retrieves and incorporates this dictionary during the data generation process.

\textbf{Human Annotations.} To evaluate data generation pipelines and trained on-device translators, we invite annotators and build \textsc{LiveChatBench}, a high-quality benchmark of 1,000 chat instances, each paired with translation outputs and the necessary translation-relevant knowledge (Table \ref{tab:sample_livechatbench}).

\textbf{Validation.} As shown in Table \ref{tab:eval_pipe}, LiveChatBench offers the advantage of enabling us to verify whether the currently constructed pipeline is effectively injecting knowledge. In this paper, we employed BM25 and an LLM-based entity extractor for keyword-based knowledge injection.

\begin{table*}[t!]
\centering
\small
\resizebox{\textwidth}{!}{%
\begin{tabular}{llll}
\toprule
\textbf{} &
  \textbf{CPU} &
  \textbf{GPU} &
  \textbf{RAM} \\
  \midrule
  
\makecell[l]{\textbf{iPhone 11 Pro}} &
  \makecell[l]{2-core Apple Lightning 2.67 GHz,\\
  4-core Apple Thunder 1.73 GHz} &
  4-core Apple 3rd-generation design GPU architecture, 0,000 MHz &
  4 GB LPDDR4X SDRAM \\
  \midrule
  
\makecell[l]{\textbf{iPhone 14 Pro Max}} &
  \makecell[l]{2-core Apple Everest 3.46 GHz,\\
  4-core Apple Sawtooth 2.02 GHz} &
  5-core Apple G14, 1,398 MHz &
  6 GB LPDDR5 SDRAM \\
  \midrule
  
\textbf{iPhone 16 Pro Max} &
  \makecell[l]{Dual-core Apple Everest (3rd generation) 4.04 GHz,\\
  Quad-core Apple Sawtooth (3rd generation) 2.40 GHz} &
  6-core Apple G17, 1,470 MHz &
  8 GB LPDDR5X SDRAM \\
  \midrule
  
\textbf{Galaxy S24+} &
  \makecell[l]{1 × ARM Cortex‑X4 3.21 GHz,\\
  2 × ARM Cortex‑A720 2.90 GHz,\\
  3 × ARM Cortex‑A720 2.60 GHz,\\
  4 × ARM Cortex‑A520 at 1.95 GHz\\} &
  Samsung Xclipse 940 1.1 GHz &
  12 GB LPDDR5X SDRAM 8,533 MT/s \\
  \midrule
  
\textbf{Galaxy S25} &
  \makecell[l]{2 × Oryon-L (2nd Gen) 4.47 GHz,\\
  6 × Oryon-M (2nd Gen) 3.53 GHz} &
  Qualcomm Adreno 830 (1.2 GHz) &
  12 GB LPDDR5X SDRAM 9,600 MT/s \\
  \bottomrule
\end{tabular}%
}
\caption{\textbf{Specifications of the mobile devices used in our experiments, including three iOS devices and two Android devices.} On iOS devices, we used Apple’s Metal framework for GPU-accelerated inference.}
\label{tab:devices}
\end{table*}

\begin{table*}[t!]
\centering
\small
% \resizebox{\textwidth}{!}{%
\resizebox{0.75\textwidth}{!}{%
\begin{tabular}{lllll}
\toprule
\textbf{} &
  \textbf{Model Weights} &
  \textbf{KV Cache} &
  \textbf{Compute Buffer} &
  \textbf{Total} \\
  \midrule
  
\makecell[l]{\textbf{Gemma-3-270M}} &
  511 MB &
  192 MB &
  18 MB &
  721 MB\\
  \midrule
  
\makecell[l]{\textbf{Qwen3-0.6B}} &
  1137 MB &
  112 MB &
  112 MB &
  1361 MB\\
  \midrule
  
\textbf{Gemma-3-1B} &
  1907 MB &
  26 MB &
  192 MB &
  2125 MB\\
  \bottomrule
\end{tabular}%
}
\caption{\textbf{Component-wise memory breakdown.} We report the total memory footprint of the Gemma-3-270M, QWEN3-0.6B, and Gemma-3-1B models, decomposed into model weights, KV cache, and compute buffers.}
\label{tab:memory_breakdown}
\end{table*}

\section{Experimental Setup}

All experiments were conducted using five smartphones: iPhone 11 Pro, iPhone 14 Pro Max, iPhone 16 Pro Max, Samsung Galaxy S24+, and Samsung Galaxy S25. We leveraged GPT-5.1 to construct the training dataset. An NVIDIA H100 NVL GPU was used to train the on-device models.

\subsection{Mobile Device Specifications}
Table \ref{tab:devices} shows the specifications of the five mobile devices used in the experiment. For iOS deployments, we utilize Apple’s Metal API for GPU acceleration through llama.cpp’s Metal backend\footnote{\url{https://github.com/mybigday/llama.rn}}. We did not run GPU experiments on the iPhone 11 Pro because the llama.cpp Metal backend requires SIMD-group operations that are available only on devices supporting Apple GPU Family 7 or later (as specified in Apple’s Metal feature set tables), which the iPhone 11 Pro does not support.

\subsection{Baselines}

To investigate whether on-device AI can be applied to real-world applications, we select four models: MLKit \cite{mlkit}, Gemma-3-270M \cite{gemmateam2025gemma3technicalreport}, Qwen3-0.6B \cite{yang2025qwen3technicalreport}, and Gemma-3-1B \cite{gemmateam2025gemma3technicalreport}. We also choose two models to measure the gap between commercial models and on-device models: Google Translate API\footnote{\url{https://pypi.org/project/googletrans/}} and GPT-5.1 \cite{OpenAI_GPT5.1}.

\begin{figure*}[t]
    \centering
    % (e.g., \textit{Are your living and working spaces clean and organized?}). 
    % \vspace{-10pt}
    \includegraphics[width=\textwidth]{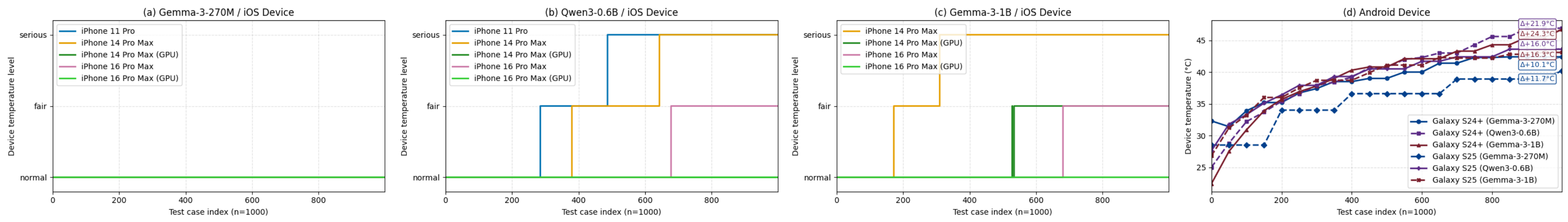}
    \caption{Mobile device temperature results on LiveChatBench across different on-device models.}
    \label{fig:results_temperature}
    % \vspace{-13pt}
\end{figure*}

\begin{figure*}[t]
    \centering
    % (e.g., \textit{Are your living and working spaces clean and organized?}). 
    % \vspace{-10pt}
    \includegraphics[width=\textwidth]{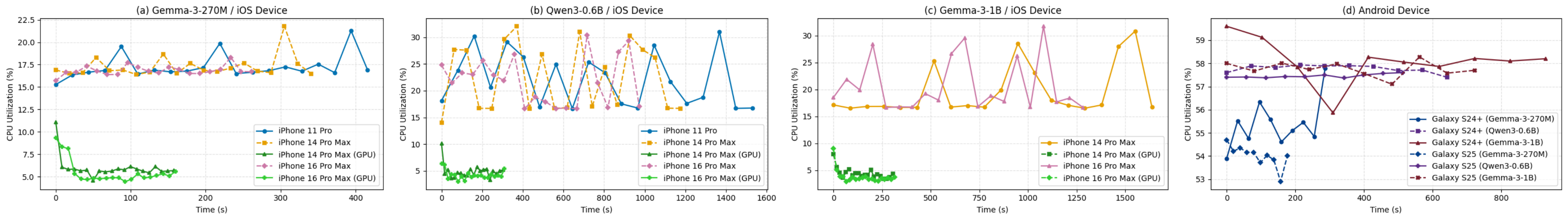}
    \caption{CPU utilization results on LiveChatBench across different on-device models.}
    \label{fig:results_cpu_util}
    % \vspace{-13pt}
\end{figure*}

\begin{table}[t]
\centering
\footnotesize
\resizebox{0.3\textwidth}{!}{
\begin{tabular}{@{}lll@{}}
\toprule
\textbf{Parameter} & \textbf{Value} \\ \midrule
\texttt{n\_ctx} & 512 \\ \midrule
\texttt{n\_threads} & 2 \\ \midrule
\texttt{n\_batch} & 256 \\ \midrule
\texttt{n\_uBatch} & 192 \\ \midrule
\texttt{flash\_attn\_type} & \texttt{On} \\ \midrule
\texttt{n\_predict} & 128 \\ \midrule
\texttt{temperature} & 0.01 \\ \midrule
\texttt{top\_p} & 0.95 \\
\bottomrule
\end{tabular}}
\caption{\textbf{Model and completion parameters.} We employ a unified configuration for all on-device models.}
\label{tab:inference_config}
\end{table}

\subsection{Implementation Details}

\textbf{Training Configuration.} This section details the experimental setup and the hyperparameter settings used throughout this work. Table \ref{tab:training_config} summarizes the common training parameters that are uniformly applied across all on-device models and datasets.

\textbf{Inference.} To improve efficiency, we use a simple prompt for inference: \textit{"Translate the following sentence into English. Only include the translated result, do not explain the result."}. Table \ref{tab:inference_config} shows the model parameters and completion parameters used for inference on each mobile device.

\subsection{Evaluation Protocols}

% ⚠️ Figure Appendix에 추가
We evaluate translation performance on three Korean–English benchmarks: LiveChatBench (1,000 parallel sentence pairs; hereafter, parallel pairs), FLORES-200 (1,012 parallel pairs) \cite{nllbteam2022languageleftbehindscaling}, and WMT24++ (998 parallel pairs) \cite{deutsch-etal-2025-wmt24}. We report BLEU~\cite{10.3115/1073083.1073135}, ChrF++~\cite{popovic-2015-chrf}, and use GPT-5.1 to assess six translation error types via the Focus Sentence Prompting format~\cite{domhan-zhu-2025-evaluation}, following~\cite{li-etal-2025-mqm} (Figure \ref{fig:eval_trans_prompt}).

\begin{table}[t]
\centering
\footnotesize
\resizebox{0.49\textwidth}{!}{%
\begin{tabular}{|l|r|r|}
\hline
\cellcolor{ForestGreen} \textbf{Mobile Device} & \cellcolor{ForestGreen} \textbf{TTFT (ms)} & \cellcolor{ForestGreen} \textbf{Runtime (s)} \\ \hline
\multicolumn{3}{|c|}{\cellcolor{lightgray} \textbf{Gemma-3-270M-IT\textsuperscript{\dag}}} \\ \hline
\textbf{iPhone 11 Pro} & 140.9460 & 0.4161 \\ \hline
\textbf{iPhone 14 Pro Max} & 121.4954 & 0.3412 \\ \hline
\textbf{iPhone 14 Pro Max\textsuperscript{$\ddagger$}} & 74.7435 & 0.1582 \\ \hline
% \textbf{iPhone 14 Pro Max\textsuperscript{$\ddagger$}} & 74.7435 & 0.5834 \\ \hline
\textbf{iPhone 16 Pro Max} & 79.3681 & 0.2608 \\ \hline
\textbf{iPhone 16 Pro Max\textsuperscript{$\ddagger$}} & 63.3153 & 0.1598 \\ \hline
% \textbf{iPhone 16 Pro Max\textsuperscript{$\ddagger$}} & 63.3153 & 0.5878 \\ \hline
\textbf{Samsung Galaxy S24+} & 48.0430 & 0.2868 \\ \hline
\textbf{Samsung Galaxy S25}& 38.4525 & 0.1769 \\ \hline
\multicolumn{3}{|c|}{\cellcolor{lightgray} \textbf{Qwen3-0.6B\textsuperscript{\dag}}} \\ \hline
\textbf{iPhone 11 Pro} & 811.4484 & 1.5294 \\ \hline
\textbf{iPhone 14 Pro Max} & 637.4881 & 1.1753 \\ \hline
\textbf{iPhone 14 Pro Max\textsuperscript{$\ddagger$}} & 158.7082 & 0.3022 \\ \hline
% \textbf{iPhone 14 Pro Max\textsuperscript{$\ddagger$}} & 158.7082 & 1.1157 \\ \hline
\textbf{iPhone 16 Pro Max} & 460.8820 & 0.9730 \\ \hline
\textbf{iPhone 16 Pro Max\textsuperscript{$\ddagger$}} & 149.2322 & 0.3104 \\ \hline
% \textbf{iPhone 16 Pro Max\textsuperscript{$\ddagger$}} & 149.2322 & 1.1458 \\ \hline
\textbf{Samsung Galaxy S24+} & 186.3310 & 0.6413 \\ \hline
\textbf{Samsung Galaxy S25}& 189.9350 & 0.5126 \\ \hline
\multicolumn{3}{|c|}{\cellcolor{lightgray} \textbf{Gemma-3-1B-IT\textsuperscript{\dag}}} \\ \hline
\textbf{iPhone 14 Pro Max} & 784.0577 & 1.6371 \\ \hline
\textbf{iPhone 14 Pro Max\textsuperscript{$\ddagger$}} & 159.3810 & 0.3046 \\ \hline
% \textbf{iPhone 14 Pro Max\textsuperscript{$\ddagger$}} & 159.3810 & 1.1192 \\ \hline
\textbf{iPhone 16 Pro Max} & 536.1123 & 1.2821 \\ \hline
\textbf{iPhone 16 Pro Max\textsuperscript{$\ddagger$}} & 151.2176 & 0.3126 \\ \hline
% \textbf{iPhone 16 Pro Max\textsuperscript{$\ddagger$}} & 151.2176 & 1.1483 \\ \hline
\textbf{Samsung Galaxy S24+} & 230.3490 & 0.9283 \\ \hline
\textbf{Samsung Galaxy S25}& 240.3030 & 0.7219 \\ \hline
\end{tabular}
}
\caption{\textbf{Evaluation of three on-device models on LiveChatBench across five mobile devices.} \textsuperscript{\dag} denotes models trained on the dataset constructed in Section \ref{sec:datasets}. \textsuperscript{$\ddagger$} indicates results measured in an environment where the mobile device’s GPU was used.} 
\label{tab:result_ttft_runtime}
\end{table}

\section{On-Device Performance Results}
\label{sec:on-device-results}

\textbf{GPU acceleration is a key enabler for real-world on-device applications.} We evaluate the thermal state and CPU utilization of three iOS devices and two Android devices on LiveChatBench. Since Apple does not expose numerical device temperatures to iOS developers, we report the thermal states of iOS devices using Apple's four-level classification. The \textit{nominal} state corresponds to temperatures within normal limits, while \textit{fair} denotes slightly elevated temperatures. The \textit{serious} and \textit{critical} states both reflect high-temperature conditions that significantly affect overall system performance and require the device to cool down immediately.

Figures \ref{fig:results_temperature}(a)-(c) show that thermal conditions remain stable across devices when GPU acceleration is enabled. As shown in Figures \ref{fig:results_cpu_util}(a)-(c), CPU utilization, an important metric for real-world applications, remains around 5\% with minimal fluctuation under GPU acceleration. These results highlight the importance of considering users' device environments when deploying on-device models as a service, as some devices do not support GPU acceleration (e.g., the iPhone 11 Pro in Table \ref{tab:devices}). Although the Galaxy S24+ and Galaxy S25 achieve fast average runtimes (Table \ref{tab:result_ttft_runtime}), their temperatures rise above 45\textdegree C and CPU utilization approaches 60\% under CPU-only execution, as shown in Figure \ref{fig:results_temperature}(d) and Figure \ref{fig:results_cpu_util}(d), respectively. These thermal and computational demands may hinder the practical deployment of on-device models in real-world services without GPU acceleration.

\textbf{On-device model selection is significantly constrained in mobile environments.} Table \ref{tab:result_ttft_runtime} shows the average time to first token (TTFT) and runtime on LiveChatBench for three on-device models across three iOS devices and two Android devices. We observe that, in the GPU environments of the devices, trained Gemma-3-270M exhibits an average runtime that is approximately 1.9178× faster and a average TTFT that is approximately 2.2503× faster than those of Qwen3-0.6B and Gemma-3-1B, demonstrating markedly better suitability for rapidly translating users’ chat messages.

Considering that a fully charged iPhone 16 Pro Max stores 65,376 J of energy, Gemma-3-270M and Gemma-3-1B, which consume 0.027 J and 0.1 J per token respectively, can generate up to 2,421,333 and 653,760 tokens, corresponding to a factor of approximately 3.703× \cite{6237004,10.1109/ISCA.2016.30}. Moreover, the iPhone 11 Pro cannot run a 1B-parameter model due to its limited memory capacity (Table \ref{tab:memory_breakdown}). Therefore, to accommodate users with diverse devices, task clarification, the specifications of devices, and the selection of model size must be carefully considered.

\begin{table*}[t]
\centering
\footnotesize
\resizebox{\textwidth}{!}{
\begin{tabular}{ccccccccccc}
\toprule
\multirow{2}{*}{\textbf{Method}} & \multicolumn{3}{c}{\textbf{LiveChatBench}} & \multicolumn{3}{c}{\textbf{FLORES-200}} & \multicolumn{3}{c}{\textbf{WMT24++}} \\
\cmidrule(lr){2-4} \cmidrule(lr){5-7} \cmidrule(lr){8-10}
& \textbf{BLEU ($\uparrow$)} & \textbf{ChrF++ ($\uparrow$)} & \textbf{FSP ($\uparrow$)} & \textbf{BLEU ($\uparrow$)} & \textbf{ChrF++ ($\uparrow$)} & \textbf{FSP ($\uparrow$)} & \textbf{BLEU ($\uparrow$)} & \textbf{ChrF++ ($\uparrow$)} & \textbf{FSP ($\uparrow$)}\\
\midrule
MLKit & 0.0489 & 17.4392 & 28.2600 & 0.2556 & 44.9525 & 52.5267 & 0.1737 & 39.5039 & 50.3096\\
Google Translate API & 0.1678 & 34.2974 & 59.7250 & \textbf{0.4449} & \textbf{60.5056} & \underline{94.1294} & \textbf{0.3333} & \textbf{53.5668} & \underline{91.7806} \\
GPT-5.1 & 0.2679 & 45.2609 & \textbf{70.2210} & \underline{0.4113} & \underline{59.0201} & \textbf{96.8607} & \underline{0.2998} & \underline{50.8218} & \textbf{96.2745}\\

\cmidrule(lr){1-10}
Gemma-3-270M-IT & 0.0097 & 5.3872 & 15.1300 & 0.0740 & 20.9606 & 23.8636 & 0.0553 & 17.4603 & 23.1293\\
Qwen3-0.6B & 0.0017 & 8.6502 & 5.9980 & 0.0261 & 21.9348 & 4.8715 & 0.0221 & 18.2811 & 5.7615\\
Gemma-3-1B-IT & 0.0430 & 17.5645 & 35.2680 & 0.2301  & 42.4403 & 65.5188 & 0.1435 & 33.4784 & 59.9639\\
\midrule
Gemma-3-270M-IT\textsuperscript{\dag} & \cellcolor{ForestGreen}0.2485 & \cellcolor{ForestGreen}43.9363 & \cellcolor{ForestGreen}62.9280 & \cellcolor{ForestGreen}0.1328 & \cellcolor{ForestGreen}32.0134 & \cellcolor{ForestGreen}31.7738 & \cellcolor{ForestGreen}0.1102 & \cellcolor{ForestGreen}30.5162 & \cellcolor{ForestGreen}40.7335\\
Qwen3-0.6B\textsuperscript{\dag} & \cellcolor{ForestGreen}\underline{0.2689} & \cellcolor{ForestGreen}\underline{45.7545} & \cellcolor{ForestGreen}65.9700 & \cellcolor{ForestGreen}0.1794 & \cellcolor{ForestGreen}37.7578 & \cellcolor{ForestGreen}47.2826 & \cellcolor{ForestGreen}0.1402 & \cellcolor{ForestGreen}34.4146 & \cellcolor{ForestGreen}50.4719 \\
Gemma-3-1B-IT\textsuperscript{\dag} & \cellcolor{ForestGreen}\textbf{0.2978} & \cellcolor{ForestGreen}\textbf{48.4404} & \cellcolor{ForestGreen}\underline{67.8740} & \cellcolor{ForestGreen}0.1978 & \cellcolor{ForestGreen}39.3742 & \cellcolor{ForestGreen}53.2401 & \cellcolor{ForestGreen}0.1556 & \cellcolor{ForestGreen}36.4609 & \cellcolor{ForestGreen}54.7896\\
\bottomrule
\end{tabular}
}
\caption{\textbf{Comparison of model performance across three different translation datasets.} Higher values indicate better translation quality. \textsuperscript{\dag} denotes models trained on the dataset constructed in Section \ref{sec:datasets}.}
\vspace{-10pt}
\label{tab:main_result_trans}
\end{table*}

\begin{figure}[t]
    \centering
    % (e.g., \textit{Are your living and working spaces clean and organized?}). 
    % \vspace{-10pt}
    \includegraphics[width=0.47\textwidth]{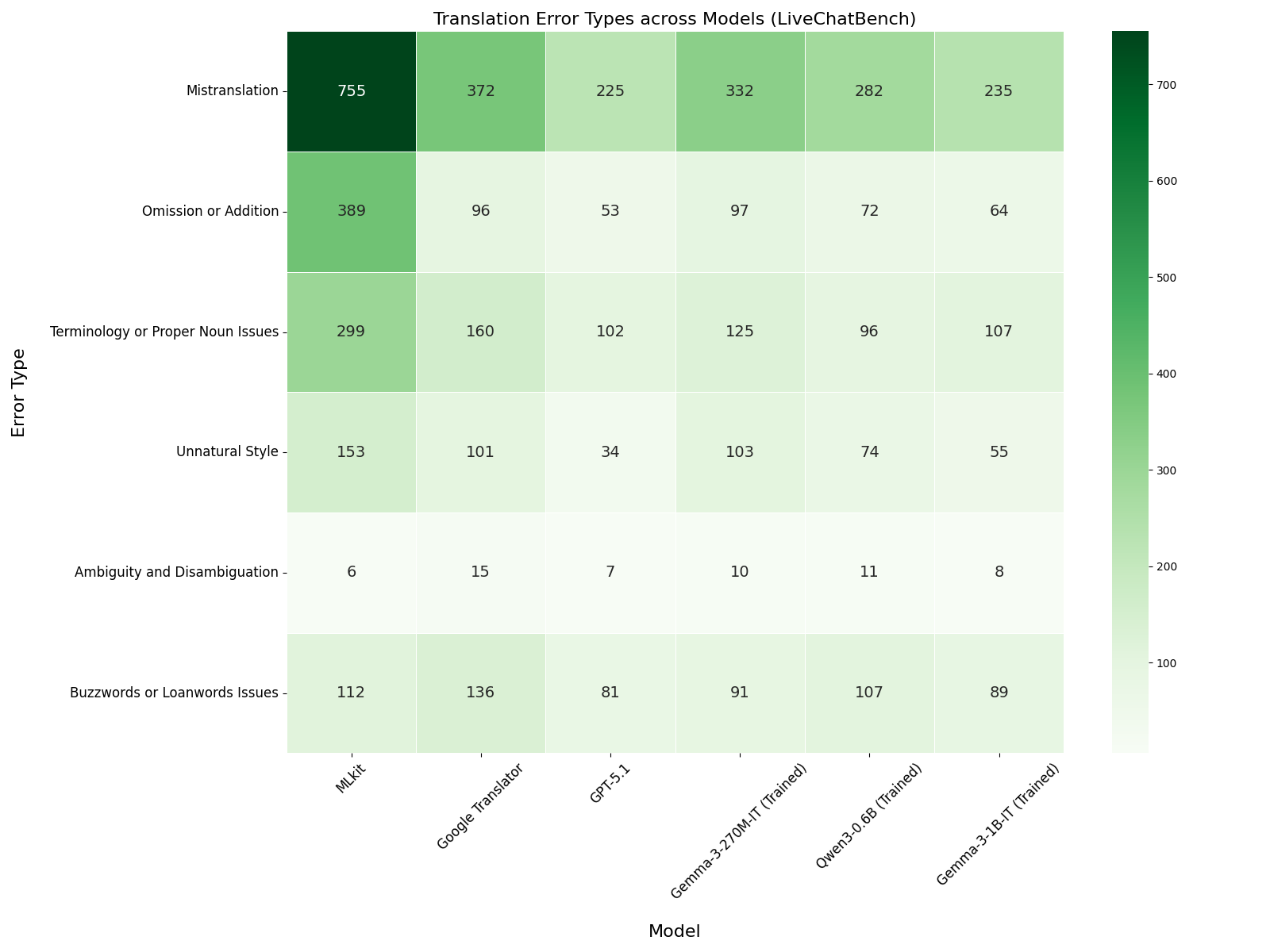}
    \caption{\textbf{Heatmap of six types of translation errors.} Darker colors indicate a higher number of errors.}
    \label{fig:heatmap_livechatbench}
    % \vspace{-13pt}
\end{figure}

\section{Translation Results}

\textbf{On-device models demonstrate excellent domain adaptation performance.} Table \ref{tab:main_result_trans} demonstrates the translation performance of each model on three Korean–English translation benchmarks. We observe that on-device models exhibit a substantial performance improvement after training. Indeed, on LiveChatBench, the Qwen3-0.6B shows 158×, 5.29×, and 11× improvements in BLEU, ChrF++, and FSP, respectively. It demonstrates the advantages and potential of on-device models that, with limited resources, can be deployed in real-world applications for well-targeted domain-specific tasks.

\textbf{When trained for the specific task, on-device models achieve performance comparable to that of commercially available models.} In Table \ref{tab:main_result_trans}, We find that, on LiveChatBench, the three on-device models exhibit performance comparable to GPT-5.1 despite their small model sizes. Meanwhile, Figure \ref{fig:heatmap_livechatbench} shows the number of errors recorded for each of the six translation error types only when the severity score exceeded 50. We also observe that, for the six types of translation errors, all three domain-adapted models outperform the Google Translate API and achieve performance close to that of GPT-5.1. However, the results presented in Table \ref{tab:main_result_trans}, Figure \ref{fig:heatmap_flores200}, and Figure \ref{fig:heatmap_wmtpp} indicate that, on FLORES-200 and WMT24++, the performance of on-device models still falls short of that of commercial models. We assume that the observed performance gap primarily stems from differences in task characteristics, particularly the average sequence length: LiveChatBench has a mean length of 13.98, whereas FLORES-200 and WMT24++ have mean lengths of 65.18 and 97.52, respectively, suggesting that on-device AI still has room for improvement.

\begin{figure}[t]
    \centering
    % (e.g., \textit{Are your living and working spaces clean and organized?}). 
    % \vspace{-10pt}
    \includegraphics[width=0.47\textwidth]{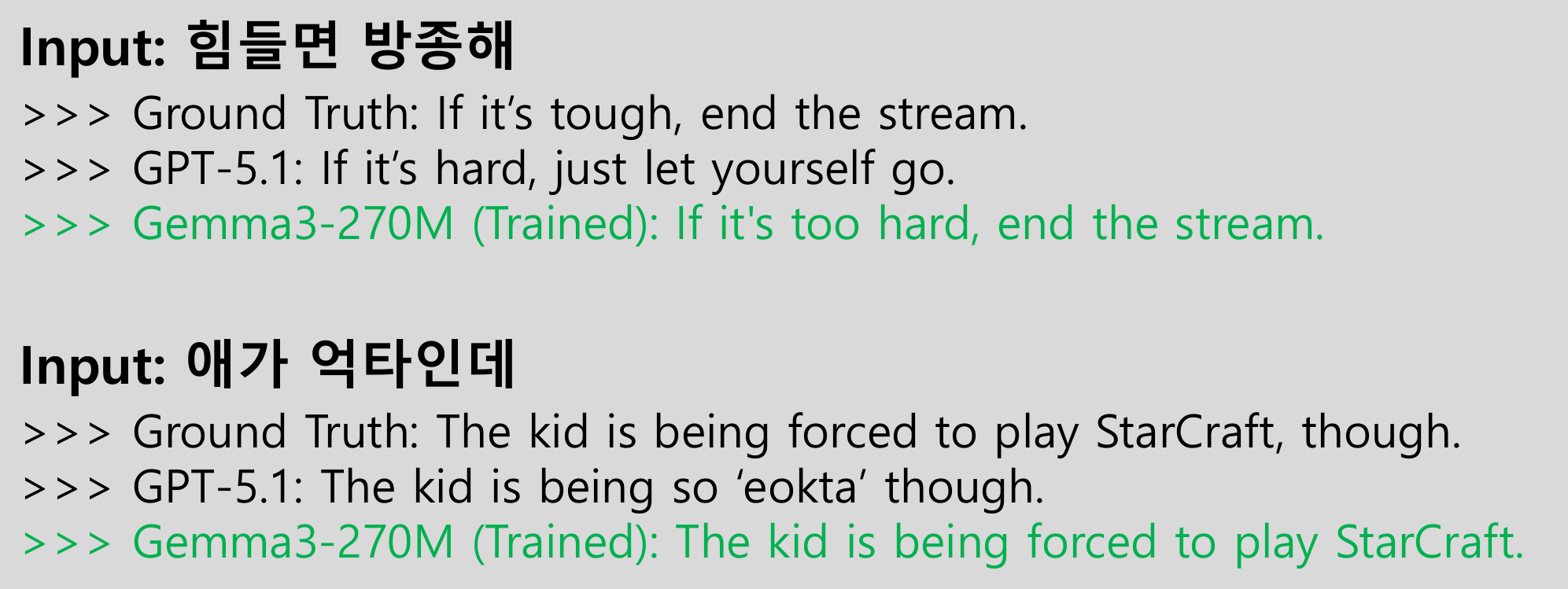}
    \caption{Comparison of the translation outputs of GPT-5.1 and trained Gemma-3-270M.}
    \label{fig:qualitative_analysis}
    % \vspace{-13pt}
\end{figure}

\textbf{Qualitative Analysis.} Figure \ref{fig:qualitative_analysis} illustrates how domain adaptation helps compact models interpret live-streaming terminology. GPT-5.1 mistranslates “방종” as “let yourself go” and romanizes “억타” as “eokta,” whereas the adapted Gemma-3-270M correctly conveys “end the stream” and “being forced to play StarCraft,” respectively. These examples suggest that targeted adaptation can enable compact models to capture domain-specific meanings missed by a general-purpose commercial model, despite their limited capacity.

% shows the translation results of LiveChatBench samples from a commercial model and an on-device model. We find that on-device models can be highly efficient when applied to well-targeted real-world tasks. 

\section{Conclusion}

In summary, this paper provides empirical evidence on the key considerations when deploying on-device models in mobile environments and on the level of performance that can be expected in practice through extensive experiments. To the best of our knowledge, this is the first practical investigation conducted across diverse mobile device environments, and we believe that our findings provide valuable insights for the on-device AI field.

\section*{Limitations}

This work has three main limitations, which future research could address. First, \textbf{Language Coverage}: Due to budgetary and other constraints, this study conducts experiments exclusively on datasets consisting of Korean–English parallel sentence pairs. Since we did not evaluate the performance of on-device models on chat data that includes meme-related terminology and slang across diverse languages, the generalizability of our findings may be limited. In future work, we should investigate the extent to which domain adaptation is effective for multilingual data.

Second, \textbf{Meme and Slang Glossary.} The collected terminology dictionary used to construct the training dataset contains 656 entries, which is insufficient to comprehensively cover all expressions that exist in real-world settings. In future work, we will need to collect a larger number of terms and focus on how to automatically gather newly emerging and disappearing terms on a daily basis and integrate them into our data construction pipeline.

Finally, \textbf{Limitations of Resource Measurement in Realistic App Execution.} Although we measured mobile-device resource utilization (Section \ref{sec:on-device-results}) during translation, the experiments primarily represent standalone translation execution. In practical settings, translation runs concurrently with application logic and OS-level tasks, which may introduce contention and variability in resource allocation (e.g., thermal throttling, background activity, UI thread workload). As a result, our measurements provide an upper-/lower-bound estimate under controlled conditions rather than a definitive characterization of resource usage during real app operation. Future work will conduct end-to-end profiling within a representative third-party livestreaming app scenario and report per-component breakdowns to improve ecological validity.

\section*{Ethical Considerations} 

This study followed established ethical guidelines to ensure the integrity and fairness of all experiments. Data collected from the livestreaming platform were used in accordance with the platform’s permissions and contained no personally identifiable information. The Korean-language datasets were used exclusively for academic purposes. All parallel sentence pairs were created through ethically appropriate procedures. The experimental design incorporated strict safeguards for data privacy and protection. All human participants were informed of the study’s purpose and procedures and were free to withdraw at any time without penalty. Overall, we sought to conduct AI research that is both scientifically rigorous and ethically responsible, with due respect for privacy, intellectual property, and participant well-being.

% \section*{Acknowledgments}

% This document has been adapted
% by Steven Bethard, Ryan Cotterell and Rui Yan
% from the instructions for earlier ACL and NAACL proceedings, including those for
% ACL 2019 by Douwe Kiela and Ivan Vuli\'{c},
% NAACL 2019 by Stephanie Lukin and Alla Roskovskaya,
% ACL 2018 by Shay Cohen, Kevin Gimpel, and Wei Lu,
% NAACL 2018 by Margaret Mitchell and Stephanie Lukin,
% Bib\TeX{} suggestions for (NA)ACL 2017/2018 from Jason Eisner,
% ACL 2017 by Dan Gildea and Min-Yen Kan,
% NAACL 2017 by Margaret Mitchell,
% ACL 2012 by Maggie Li and Michael White,
% ACL 2010 by Jing-Shin Chang and Philipp Koehn,
% ACL 2008 by Johanna D. Moore, Simone Teufel, James Allan, and Sadaoki Furui,
% ACL 2005 by Hwee Tou Ng and Kemal Oflazer,
% ACL 2002 by Eugene Charniak and Dekang Lin,
% and earlier ACL and EACL formats written by several people, including
% John Chen, Henry S. Thompson and Donald Walker.
% Additional elements were taken from the formatting instructions of the \emph{International Joint Conference on Artificial Intelligence} and the \emph{Conference on Computer Vision and Pattern Recognition}.

% Bibliography entries for the entire Anthology, followed by custom entries
%\bibliography{anthology,custom}
% Custom bibliography entries only
\bibliography{custom}

@misc{nllbteam2022languageleftbehindscaling,
      title={No Language Left Behind: Scaling Human-Centered Machine Translation}, 
      author={NLLB Team and Marta R. Costa-jussà and James Cross and Onur Çelebi and Maha Elbayad and Kenneth Heafield and Kevin Heffernan and Elahe Kalbassi and Janice Lam and Daniel Licht and Jean Maillard and Anna Sun and Skyler Wang and Guillaume Wenzek and Al Youngblood and Bapi Akula and Loic Barrault and Gabriel Mejia Gonzalez and Prangthip Hansanti and John Hoffman and Semarley Jarrett and Kaushik Ram Sadagopan and Dirk Rowe and Shannon Spruit and Chau Tran and Pierre Andrews and Necip Fazil Ayan and Shruti Bhosale and Sergey Edunov and Angela Fan and Cynthia Gao and Vedanuj Goswami and Francisco Guzmán and Philipp Koehn and Alexandre Mourachko and Christophe Ropers and Safiyyah Saleem and Holger Schwenk and Jeff Wang},
      year={2022},
      eprint={2207.04672},
      archivePrefix={arXiv},
      primaryClass={cs.CL},
      url={https://arxiv.org/abs/2207.04672}, 
}

@inproceedings{deutsch-etal-2025-wmt24,
    title = "{WMT}24++: Expanding the Language Coverage of {WMT}24 to 55 Languages {\&} Dialects",
    author = "Deutsch, Daniel  and
      Briakou, Eleftheria  and
      Caswell, Isaac Rayburn  and
      Finkelstein, Mara  and
      Galor, Rebecca  and
      Juraska, Juraj  and
      Kovacs, Geza  and
      Lui, Alison  and
      Rei, Ricardo  and
      Riesa, Jason  and
      Rijhwani, Shruti  and
      Riley, Parker  and
      Salesky, Elizabeth  and
      Trabelsi, Firas  and
      Winkler, Stephanie  and
      Zhang, Biao  and
      Freitag, Markus",
    editor = "Che, Wanxiang  and
      Nabende, Joyce  and
      Shutova, Ekaterina  and
      Pilehvar, Mohammad Taher",
    booktitle = "Findings of the Association for Computational Linguistics: ACL 2025",
    month = jul,
    year = "2025",
    address = "Vienna, Austria",
    publisher = "Association for Computational Linguistics",
    url = "https://aclanthology.org/2025.findings-acl.634/",
    doi = "10.18653/v1/2025.findings-acl.634",
    pages = "12257--12284",
    ISBN = "979-8-89176-256-5"
}

@inproceedings{domhan-zhu-2025-evaluation,
    title = "Same evaluation, more tokens: On the effect of input length for machine translation evaluation using Large Language Models",
    author = "Domhan, Tobias  and
      Zhu, Dawei",
    editor = "Christodoulopoulos, Christos  and
      Chakraborty, Tanmoy  and
      Rose, Carolyn  and
      Peng, Violet",
    booktitle = "Proceedings of the 2025 Conference on Empirical Methods in Natural Language Processing",
    month = nov,
    year = "2025",
    address = "Suzhou, China",
    publisher = "Association for Computational Linguistics",
    url = "https://aclanthology.org/2025.emnlp-main.402/",
    doi = "10.18653/v1/2025.emnlp-main.402",
    pages = "7940--7958",
    ISBN = "979-8-89176-332-6"
}

@inproceedings{sun-etal-2024-toward,
    title = "Toward Informal Language Processing: Knowledge of Slang in Large Language Models",
    author = "Sun, Zhewei  and
      Hu, Qian  and
      Gupta, Rahul  and
      Zemel, Richard  and
      Xu, Yang",
    editor = "Duh, Kevin  and
      Gomez, Helena  and
      Bethard, Steven",
    booktitle = "Proceedings of the 2024 Conference of the North American Chapter of the Association for Computational Linguistics: Human Language Technologies (Volume 1: Long Papers)",
    month = jun,
    year = "2024",
    address = "Mexico City, Mexico",
    publisher = "Association for Computational Linguistics",
    url = "https://aclanthology.org/2024.naacl-long.94/",
    doi = "10.18653/v1/2024.naacl-long.94",
    pages = "1683--1701"
}

@inproceedings{sun-etal-2022-semantically,
    title = "Semantically Informed Slang Interpretation",
    author = "Sun, Zhewei  and
      Zemel, Richard  and
      Xu, Yang",
    editor = "Carpuat, Marine  and
      de Marneffe, Marie-Catherine  and
      Meza Ruiz, Ivan Vladimir",
    booktitle = "Proceedings of the 2022 Conference of the North American Chapter of the Association for Computational Linguistics: Human Language Technologies",
    month = jul,
    year = "2022",
    address = "Seattle, United States",
    publisher = "Association for Computational Linguistics",
    url = "https://aclanthology.org/2022.naacl-main.383/",
    doi = "10.18653/v1/2022.naacl-main.383",
    pages = "5213--5231"
}

@inproceedings{wuraola-etal-2024-understanding,
    title = "Understanding Slang with {LLM}s: Modelling Cross-Cultural Nuances through Paraphrasing",
    author = "Wuraola, Ifeoluwa  and
      Dethlefs, Nina  and
      Marciniak, Daniel",
    editor = "Al-Onaizan, Yaser  and
      Bansal, Mohit  and
      Chen, Yun-Nung",
    booktitle = "Proceedings of the 2024 Conference on Empirical Methods in Natural Language Processing",
    month = nov,
    year = "2024",
    address = "Miami, Florida, USA",
    publisher = "Association for Computational Linguistics",
    url = "https://aclanthology.org/2024.emnlp-main.869/",
    doi = "10.18653/v1/2024.emnlp-main.869",
    pages = "15525--15531"
}

@inproceedings{de-gibert-etal-2025-scaling,
    title = "Scaling Low-Resource {MT} via Synthetic Data Generation with {LLM}s",
    author = {de Gibert, Ona  and
      Attieh, Joseph  and
      Vahtola, Teemu  and
      Aulamo, Mikko  and
      Li, Zihao  and
      V{\'a}zquez, Ra{\'u}l  and
      Hu, Tiancheng  and
      Tiedemann, J{\"o}rg},
    editor = "Christodoulopoulos, Christos  and
      Chakraborty, Tanmoy  and
      Rose, Carolyn  and
      Peng, Violet",
    booktitle = "Proceedings of the 2025 Conference on Empirical Methods in Natural Language Processing",
    month = nov,
    year = "2025",
    address = "Suzhou, China",
    publisher = "Association for Computational Linguistics",
    url = "https://aclanthology.org/2025.emnlp-main.1408/",
    doi = "10.18653/v1/2025.emnlp-main.1408",
    pages = "27662--27680",
    ISBN = "979-8-89176-332-6"
}

@inproceedings{kartik-etal-2024-synthetic,
    title = "Synthetic Data Generation and Joint Learning for Robust Code-Mixed Translation",
    author = "Kartik  and
      Soni, Sanjana  and
      Kunchukuttan, Anoop  and
      Chakraborty, Tanmoy  and
      Akhtar, Md Shad",
    editor = "Calzolari, Nicoletta  and
      Kan, Min-Yen  and
      Hoste, Veronique  and
      Lenci, Alessandro  and
      Sakti, Sakriani  and
      Xue, Nianwen",
    booktitle = "Proceedings of the 2024 Joint International Conference on Computational Linguistics, Language Resources and Evaluation (LREC-COLING 2024)",
    month = may,
    year = "2024",
    address = "Torino, Italia",
    publisher = "ELRA and ICCL",
    url = "https://aclanthology.org/2024.lrec-main.1345/",
    pages = "15480--15492"
}

@misc{yang2025qwen3technicalreport,
      title={Qwen3 Technical Report}, 
      author={An Yang and Anfeng Li and Baosong Yang and Beichen Zhang and Binyuan Hui and Bo Zheng and Bowen Yu and Chang Gao and Chengen Huang and Chenxu Lv and Chujie Zheng and Dayiheng Liu and Fan Zhou and Fei Huang and Feng Hu and Hao Ge and Haoran Wei and Huan Lin and Jialong Tang and Jian Yang and Jianhong Tu and Jianwei Zhang and Jianxin Yang and Jiaxi Yang and Jing Zhou and Jingren Zhou and Junyang Lin and Kai Dang and Keqin Bao and Kexin Yang and Le Yu and Lianghao Deng and Mei Li and Mingfeng Xue and Mingze Li and Pei Zhang and Peng Wang and Qin Zhu and Rui Men and Ruize Gao and Shixuan Liu and Shuang Luo and Tianhao Li and Tianyi Tang and Wenbiao Yin and Xingzhang Ren and Xinyu Wang and Xinyu Zhang and Xuancheng Ren and Yang Fan and Yang Su and Yichang Zhang and Yinger Zhang and Yu Wan and Yuqiong Liu and Zekun Wang and Zeyu Cui and Zhenru Zhang and Zhipeng Zhou and Zihan Qiu},
      year={2025},
      eprint={2505.09388},
      archivePrefix={arXiv},
      primaryClass={cs.CL},
      url={https://arxiv.org/abs/2505.09388}, 
}

@misc{gemmateam2025gemma3technicalreport,
      title={Gemma 3 Technical Report}, 
      author={Gemma Team and Aishwarya Kamath and Johan Ferret and Shreya Pathak and Nino Vieillard and Ramona Merhej and Sarah Perrin and Tatiana Matejovicova and Alexandre Ramé and Morgane Rivière and Louis Rouillard and Thomas Mesnard and Geoffrey Cideron and Jean-bastien Grill and Sabela Ramos and Edouard Yvinec and Michelle Casbon and Etienne Pot and Ivo Penchev and Gaël Liu and Francesco Visin and Kathleen Kenealy and Lucas Beyer and Xiaohai Zhai and Anton Tsitsulin and Robert Busa-Fekete and Alex Feng and Noveen Sachdeva and Benjamin Coleman and Yi Gao and Basil Mustafa and Iain Barr and Emilio Parisotto and David Tian and Matan Eyal and Colin Cherry and Jan-Thorsten Peter and Danila Sinopalnikov and Surya Bhupatiraju and Rishabh Agarwal and Mehran Kazemi and Dan Malkin and Ravin Kumar and David Vilar and Idan Brusilovsky and Jiaming Luo and Andreas Steiner and Abe Friesen and Abhanshu Sharma and Abheesht Sharma and Adi Mayrav Gilady and Adrian Goedeckemeyer and Alaa Saade and Alex Feng and Alexander Kolesnikov and Alexei Bendebury and Alvin Abdagic and Amit Vadi and András György and André Susano Pinto and Anil Das and Ankur Bapna and Antoine Miech and Antoine Yang and Antonia Paterson and Ashish Shenoy and Ayan Chakrabarti and Bilal Piot and Bo Wu and Bobak Shahriari and Bryce Petrini and Charlie Chen and Charline Le Lan and Christopher A. Choquette-Choo and CJ Carey and Cormac Brick and Daniel Deutsch and Danielle Eisenbud and Dee Cattle and Derek Cheng and Dimitris Paparas and Divyashree Shivakumar Sreepathihalli and Doug Reid and Dustin Tran and Dustin Zelle and Eric Noland and Erwin Huizenga and Eugene Kharitonov and Frederick Liu and Gagik Amirkhanyan and Glenn Cameron and Hadi Hashemi and Hanna Klimczak-Plucińska and Harman Singh and Harsh Mehta and Harshal Tushar Lehri and Hussein Hazimeh and Ian Ballantyne and Idan Szpektor and Ivan Nardini and Jean Pouget-Abadie and Jetha Chan and Joe Stanton and John Wieting and Jonathan Lai and Jordi Orbay and Joseph Fernandez and Josh Newlan and Ju-yeong Ji and Jyotinder Singh and Kat Black and Kathy Yu and Kevin Hui and Kiran Vodrahalli and Klaus Greff and Linhai Qiu and Marcella Valentine and Marina Coelho and Marvin Ritter and Matt Hoffman and Matthew Watson and Mayank Chaturvedi and Michael Moynihan and Min Ma and Nabila Babar and Natasha Noy and Nathan Byrd and Nick Roy and Nikola Momchev and Nilay Chauhan and Noveen Sachdeva and Oskar Bunyan and Pankil Botarda and Paul Caron and Paul Kishan Rubenstein and Phil Culliton and Philipp Schmid and Pier Giuseppe Sessa and Pingmei Xu and Piotr Stanczyk and Pouya Tafti and Rakesh Shivanna and Renjie Wu and Renke Pan and Reza Rokni and Rob Willoughby and Rohith Vallu and Ryan Mullins and Sammy Jerome and Sara Smoot and Sertan Girgin and Shariq Iqbal and Shashir Reddy and Shruti Sheth and Siim Põder and Sijal Bhatnagar and Sindhu Raghuram Panyam and Sivan Eiger and Susan Zhang and Tianqi Liu and Trevor Yacovone and Tyler Liechty and Uday Kalra and Utku Evci and Vedant Misra and Vincent Roseberry and Vlad Feinberg and Vlad Kolesnikov and Woohyun Han and Woosuk Kwon and Xi Chen and Yinlam Chow and Yuvein Zhu and Zichuan Wei and Zoltan Egyed and Victor Cotruta and Minh Giang and Phoebe Kirk and Anand Rao and Kat Black and Nabila Babar and Jessica Lo and Erica Moreira and Luiz Gustavo Martins and Omar Sanseviero and Lucas Gonzalez and Zach Gleicher and Tris Warkentin and Vahab Mirrokni and Evan Senter and Eli Collins and Joelle Barral and Zoubin Ghahramani and Raia Hadsell and Yossi Matias and D. Sculley and Slav Petrov and Noah Fiedel and Noam Shazeer and Oriol Vinyals and Jeff Dean and Demis Hassabis and Koray Kavukcuoglu and Clement Farabet and Elena Buchatskaya and Jean-Baptiste Alayrac and Rohan Anil and Dmitry and Lepikhin and Sebastian Borgeaud and Olivier Bachem and Armand Joulin and Alek Andreev and Cassidy Hardin and Robert Dadashi and Léonard Hussenot},
      year={2025},
      eprint={2503.19786},
      archivePrefix={arXiv},
      primaryClass={cs.CL},
      url={https://arxiv.org/abs/2503.19786}, 
}

@misc{xu2024ondevicelanguagemodelscomprehensive,
      title={On-Device Language Models: A Comprehensive Review}, 
      author={Jiajun Xu and Zhiyuan Li and Wei Chen and Qun Wang and Xin Gao and Qi Cai and Ziyuan Ling},
      year={2024},
      eprint={2409.00088},
      archivePrefix={arXiv},
      primaryClass={cs.CL},
      url={https://arxiv.org/abs/2409.00088}, 
}

@inproceedings{bohdal-etal-2025-device,
    title = "On-device System of Compositional Multi-tasking in Large Language Models",
    author = "Bohdal, Ondrej  and
      Theodosiadis, Konstantinos  and
      Mpatziakas, Asterios  and
      Filippidis, Dimitrios  and
      Spyrou, Iro  and
      Zonios, Christos  and
      Drosou, Anastasios  and
      Ioannidis, Dimosthenis  and
      Lee, Kyenghun  and
      Moon, Jijoong  and
      Ko, Hyeonmok  and
      Ozay, Mete  and
      Michieli, Umberto",
    editor = "Potdar, Saloni  and
      Rojas-Barahona, Lina  and
      Montella, Sebastien",
    booktitle = "Proceedings of the 2025 Conference on Empirical Methods in Natural Language Processing: Industry Track",
    month = nov,
    year = "2025",
    address = "Suzhou (China)",
    publisher = "Association for Computational Linguistics",
    url = "https://aclanthology.org/2025.emnlp-industry.27/",
    doi = "10.18653/v1/2025.emnlp-industry.27",
    pages = "416--424",
    ISBN = "979-8-89176-333-3"
}

@misc{liu2024mobilellmoptimizingsubbillionparameter,
      title={MobileLLM: Optimizing Sub-billion Parameter Language Models for On-Device Use Cases}, 
      author={Zechun Liu and Changsheng Zhao and Forrest Iandola and Chen Lai and Yuandong Tian and Igor Fedorov and Yunyang Xiong and Ernie Chang and Yangyang Shi and Raghuraman Krishnamoorthi and Liangzhen Lai and Vikas Chandra},
      year={2024},
      eprint={2402.14905},
      archivePrefix={arXiv},
      primaryClass={cs.LG},
      url={https://arxiv.org/abs/2402.14905}, 
}

@misc{zhao2025mobilellmr1exploringlimitssubbillion,
      title={MobileLLM-R1: Exploring the Limits of Sub-Billion Language Model Reasoners with Open Training Recipes}, 
      author={Changsheng Zhao and Ernie Chang and Zechun Liu and Chia-Jung Chang and Wei Wen and Chen Lai and Sheng Cao and Yuandong Tian and Raghuraman Krishnamoorthi and Yangyang Shi and Vikas Chandra},
      year={2025},
      eprint={2509.24945},
      archivePrefix={arXiv},
      primaryClass={cs.CL},
      url={https://arxiv.org/abs/2509.24945}, 
}

@inproceedings{pham-etal-2025-slimlm,
    title = "{S}lim{LM}: An Efficient Small Language Model for On-Device Document Assistance",
    author = "Pham, Thang M.  and
      Nguyen, Phat T.  and
      Yoon, Seunghyun  and
      Lai, Viet Dac  and
      Dernoncourt, Franck  and
      Bui, Trung",
    editor = "Mishra, Pushkar  and
      Muresan, Smaranda  and
      Yu, Tao",
    booktitle = "Proceedings of the 63rd Annual Meeting of the Association for Computational Linguistics (Volume 3: System Demonstrations)",
    month = jul,
    year = "2025",
    address = "Vienna, Austria",
    publisher = "Association for Computational Linguistics",
    url = "https://aclanthology.org/2025.acl-demo.42/",
    doi = "10.18653/v1/2025.acl-demo.42",
    pages = "436--447",
    ISBN = "979-8-89176-253-4"
}

@misc{shen2025ragsynthsyntheticdatarobust,
      title={RAGSynth: Synthetic Data for Robust and Faithful RAG Component Optimization}, 
      author={Haiyang Shen and Hang Yan and Zhongshi Xing and Mugeng Liu and Yue Li and Zhiyang Chen and Yuxiang Wang and Jiuzheng Wang and Yun Ma},
      year={2025},
      eprint={2505.10989},
      archivePrefix={arXiv},
      primaryClass={cs.AI},
      url={https://arxiv.org/abs/2505.10989}, 
}

@article{mlkit,
    author  = {Google},
    title   = {Translation},
    journal = {Google for Developers},
    year    = {2019},
}

@misc{OpenAI_GPT5.1,
author = "OpenAI",
title = "GPT-5.1: A smarter, more conversational ChatGPT",
url = "https://openai.com/index/gpt-5-1/",
year = {2025}
}

@inproceedings{10.3115/1073083.1073135,
author = {Papineni, Kishore and Roukos, Salim and Ward, Todd and Zhu, Wei-Jing},
title = {BLEU: a method for automatic evaluation of machine translation},
year = {2002},
publisher = {Association for Computational Linguistics},
address = {USA},
url = {https://doi.org/10.3115/1073083.1073135},
doi = {10.3115/1073083.1073135},
booktitle = {Proceedings of the 40th Annual Meeting on Association for Computational Linguistics},
pages = {311–318},
numpages = {8},
location = {Philadelphia, Pennsylvania},
series = {ACL '02}
}

@inproceedings{popovic-2015-chrf,
    title = "chr{F}: character n-gram {F}-score for automatic {MT} evaluation",
    author = "Popovi{\'c}, Maja",
    editor = "Bojar, Ond{\v{r}}ej  and
      Chatterjee, Rajan  and
      Federmann, Christian  and
      Haddow, Barry  and
      Hokamp, Chris  and
      Huck, Matthias  and
      Logacheva, Varvara  and
      Pecina, Pavel",
    booktitle = "Proceedings of the Tenth Workshop on Statistical Machine Translation",
    month = sep,
    year = "2015",
    address = "Lisbon, Portugal",
    publisher = "Association for Computational Linguistics",
    url = "https://aclanthology.org/W15-3049/",
    doi = "10.18653/v1/W15-3049",
    pages = "392--395"
}

@inproceedings{li-etal-2025-mqm,
    title = "{MQM}-Chat: Multidimensional Quality Metrics for Chat Translation",
    author = "Li, Yunmeng  and
      Suzuki, Jun  and
      Morishita, Makoto  and
      Abe, Kaori  and
      Inui, Kentaro",
    editor = "Rambow, Owen  and
      Wanner, Leo  and
      Apidianaki, Marianna  and
      Al-Khalifa, Hend  and
      Eugenio, Barbara Di  and
      Schockaert, Steven",
    booktitle = "Proceedings of the 31st International Conference on Computational Linguistics",
    month = jan,
    year = "2025",
    address = "Abu Dhabi, UAE",
    publisher = "Association for Computational Linguistics",
    url = "https://aclanthology.org/2025.coling-main.221/",
    pages = "3283--3299"
}

@misc{allal2025smollm2smolgoesbig,
      title={SmolLM2: When Smol Goes Big -- Data-Centric Training of a Small Language Model}, 
      author={Loubna Ben Allal and Anton Lozhkov and Elie Bakouch and Gabriel Martín Blázquez and Guilherme Penedo and Lewis Tunstall and Andrés Marafioti and Hynek Kydlíček and Agustín Piqueres Lajarín and Vaibhav Srivastav and Joshua Lochner and Caleb Fahlgren and Xuan-Son Nguyen and Clémentine Fourrier and Ben Burtenshaw and Hugo Larcher and Haojun Zhao and Cyril Zakka and Mathieu Morlon and Colin Raffel and Leandro von Werra and Thomas Wolf},
      year={2025},
      eprint={2502.02737},
      archivePrefix={arXiv},
      primaryClass={cs.CL},
      url={https://arxiv.org/abs/2502.02737}, 
}

@misc{hau2025llmsmobileappspractices,
      title={LLMs in Mobile Apps: Practices, Challenges, and Opportunities}, 
      author={Kimberly Hau and Safwat Hassan and Shurui Zhou},
      year={2025},
      eprint={2502.15908},
      archivePrefix={arXiv},
      primaryClass={cs.SE},
      url={https://arxiv.org/abs/2502.15908}, 
}

@inproceedings{kim-etal-2024-click,
    title = "{CLI}c{K}: A Benchmark Dataset of Cultural and Linguistic Intelligence in {K}orean",
    author = "Kim, Eunsu  and
      Suk, Juyoung  and
      Oh, Philhoon  and
      Yoo, Haneul  and
      Thorne, James  and
      Oh, Alice",
    editor = "Calzolari, Nicoletta  and
      Kan, Min-Yen  and
      Hoste, Veronique  and
      Lenci, Alessandro  and
      Sakti, Sakriani  and
      Xue, Nianwen",
    booktitle = "Proceedings of the 2024 Joint International Conference on Computational Linguistics, Language Resources and Evaluation (LREC-COLING 2024)",
    month = may,
    year = "2024",
    address = "Torino, Italia",
    publisher = "ELRA and ICCL",
    url = "https://aclanthology.org/2024.lrec-main.296/",
    pages = "3335--3346"
}

@INPROCEEDINGS{6237004,
  author={Malladi, Krishna T. and Nothaft, Frank A. and Periyathambi, Karthika and Lee, Benjamin C. and Kozyrakis, Christos and Horowitz, Mark},
  booktitle={2012 39th Annual International Symposium on Computer Architecture (ISCA)}, 
  title={Towards energy-proportional datacenter memory with mobile DRAM}, 
  year={2012},
  volume={},
  number={},
  pages={37-48},
  doi={10.1109/ISCA.2012.6237004}}

@inproceedings{10.1109/ISCA.2016.30,
author = {Han, Song and Liu, Xingyu and Mao, Huizi and Pu, Jing and Pedram, Ardavan and Horowitz, Mark A. and Dally, William J.},
title = {EIE: efficient inference engine on compressed deep neural network},
year = {2016},
isbn = {9781467389471},
publisher = {IEEE Press},
url = {https://doi.org/10.1109/ISCA.2016.30},
doi = {10.1109/ISCA.2016.30},
booktitle = {Proceedings of the 43rd International Symposium on Computer Architecture},
pages = {243–254},
numpages = {12},
location = {Seoul, Republic of Korea},
series = {ISCA '16}
}

\clearpage
\appendix
\twocolumn

% \section{Mobile Device Specifications}
% \label{sec:appendix_device_spec}

\begin{figure}[t]
    \centering
    % (e.g., \textit{Are your living and working spaces clean and organized?}). 
    % \vspace{-10pt}
    \includegraphics[width=0.47\textwidth]{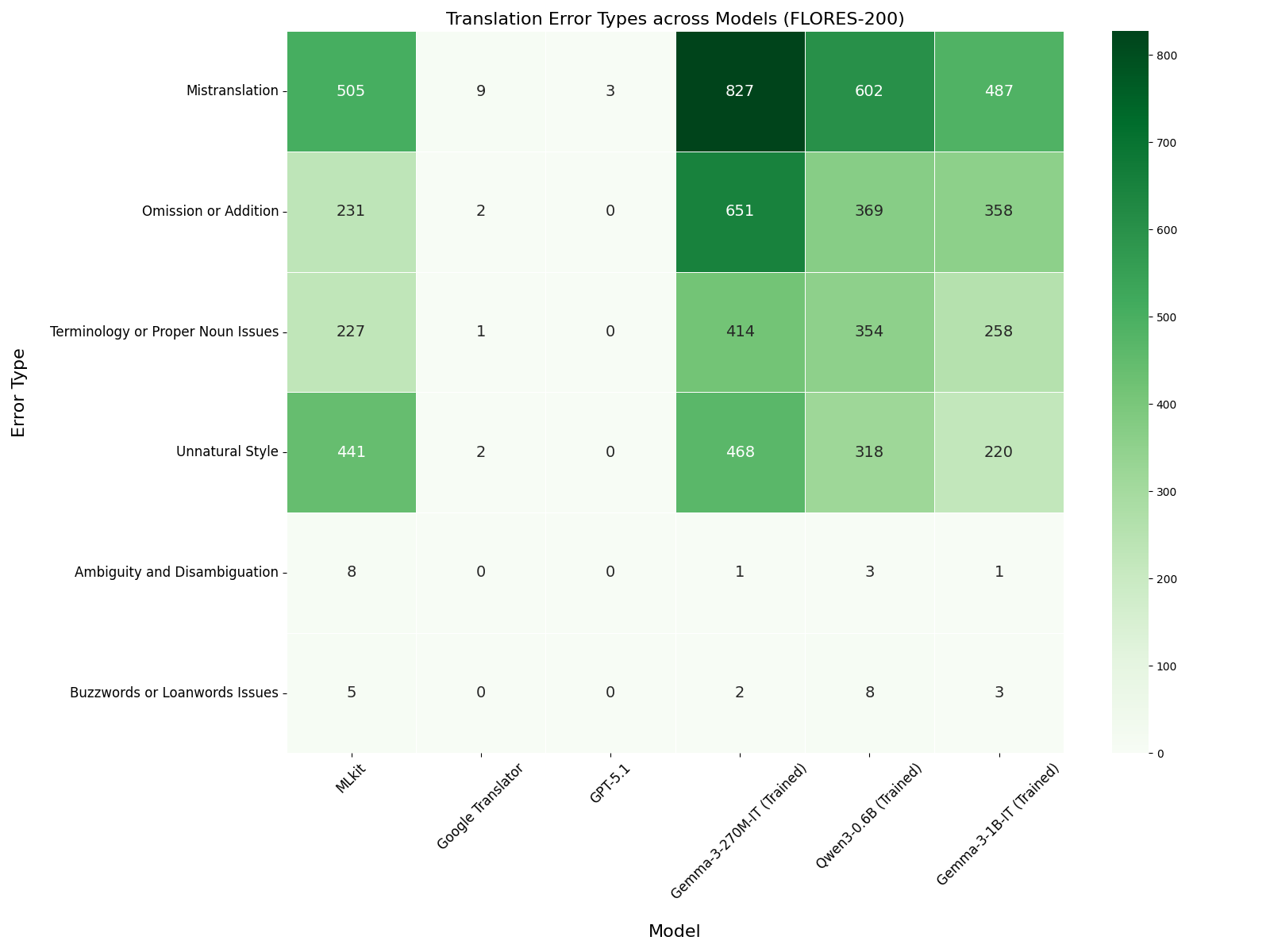}
    \caption{\textbf{Heatmap of six types of translation errors (FLORES-200).} Darker colors indicate a higher number of errors.}
    \label{fig:heatmap_flores200}
    % \vspace{-13pt}
\end{figure}

\begin{figure}[t]
    \centering
    % (e.g., \textit{Are your living and working spaces clean and organized?}). 
    % \vspace{-10pt}
    \includegraphics[width=0.47\textwidth]{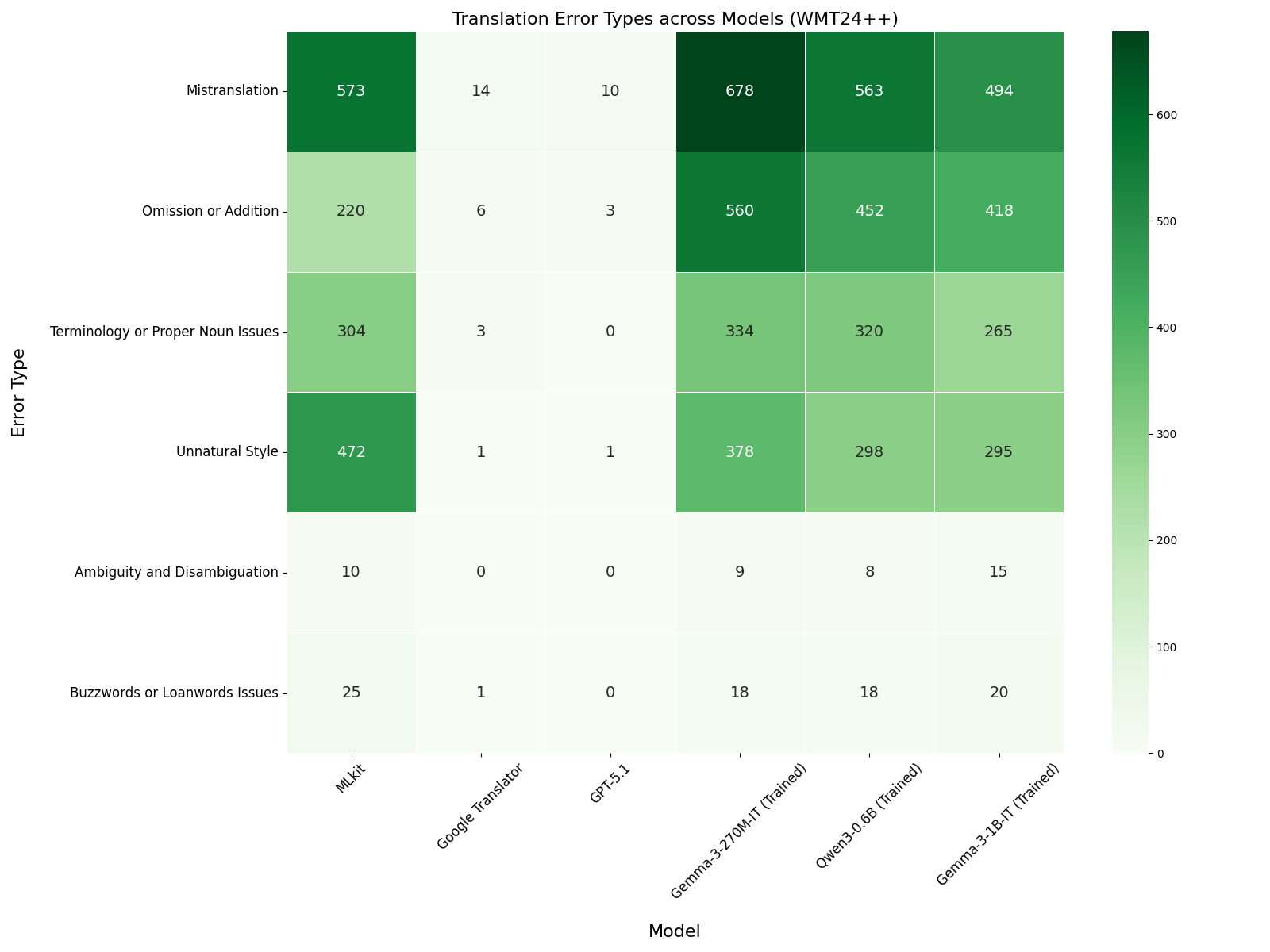}
    \caption{\textbf{Heatmap of six types of translation errors (WMT24++).} Darker colors indicate a higher number of errors.}
    \label{fig:heatmap_wmtpp}
    % \vspace{-13pt}
\end{figure}

\begin{table}[t]
\centering
\footnotesize
\resizebox{0.4\textwidth}{!}{
\begin{tabular}{@{}lll@{}}
\toprule
\textbf{Parameter} & \textbf{Value} \\ \midrule
Total Rank ($r$) & 64 \\ \midrule
Scaling Factor ($\alpha$) & 32 \\ \midrule
Target Modules & \texttt{{\{q, k, v, o, gate, up, down\}\_proj}} \\ \midrule
Optimizer & \texttt{AdamW} \\ \midrule
Warmup Ratio & 0.03 \\ \midrule
Gradient Accumulated Batch & 4 \\ \midrule
Learning Rate & 2e-4 \\ \midrule
Dropout Rate & 0.05 \\ \midrule
Max Sequence Length & 256 \\
\bottomrule
\end{tabular}}
\caption{\textbf{LoRA training hyperparameters.} We employ a unified configuration for all on-device models.}
\label{tab:training_config}
\end{table}

% Table \ref{tab:devices} shows the specifications of the five mobile devices used in the experiment. For iOS deployments, we utilize Apple’s Metal API for GPU acceleration through llama.cpp’s Metal backend\footnote{\url{https://github.com/mybigday/llama.rn}}. We did not run GPU experiments on the iPhone 11 Pro because the llama.cpp Metal backend requires SIMD-group operations that are available only on devices supporting Apple GPU Family 7 or later (as specified in Apple’s Metal feature set tables), which the iPhone 11 Pro does not support.

% \section{Implementation Details}

% \textbf{Training Configuration.} This section details the experimental setup and the hyperparameter settings used throughout this work. Table \ref{tab:training_config} summarizes the common training parameters that are uniformly applied across all on-device models and datasets.

% \textbf{Inference.} In this paper, to improve efficiency, we use a simple prompt for inference: \textit{"Translate the following sentence into English. Only include the translated result, do not explain the result."} Table \ref{tab:inference_config} shows the model parameters and completion parameters used for inference on each mobile device.

\section{Translation Error Distribution}

Figure \ref{fig:heatmap_flores200} and Figure \ref{fig:heatmap_wmtpp} show the distribution of six translation error types on the FLORES-200 and WMT24++ benchmarks. The three trained on-device models exhibit lower performance than commercial models, which is likely due to the substantial domain mismatch between the live-stream chat data used for training and the domains of the two benchmarks. Indeed, whereas LiveChatBench includes meme terminology and slang and has an average length of 13.98, FLORES-200 and WMT24++ do not, with much longer average lengths of 65.18 and 97.52, respectively. These results indicate that there is still room for improvement in on-device AI models, not only in terms of domain adaptation performance but also in achieving better generalization performance.

\clearpage

\onecolumn
\section{Prompt Description}
\label{app:prompts}
\begin{center}
\begin{tcolorbox}[width=\linewidth,colback=white, colframe=ForestGreen, title=Prompt Template (FSP) for LLM-Based Evaluation of Six Error Types]
\small
{\slshape 
\textbf{\# Role:}\\
You are an annotator for the quality of machine translation. Your task is to identify errors and assess the quality of the translation. There may be no translation errors. \\

\textbf{\# Instruction:}\\ 
Based on the source text (in <source></source> tags), machine translation surrounded (in <translation></translation> tags), and information (in <information></information> tags) identify error types in the translation and classify them. There may be no translation errors. \\

The categories of errors are:\\

- Mistranslation (Mistranslation refers to fundamental inaccuracies in the translation process, including untranslated source segments incorrect lexical choice or grammar that distorts the meaning, as well as undertranslation and overtranslation.)\\
- Omission or Addition (Missing source content (omission) or additional content not present in the source (addition) are considered to be Omission or Addition errors.)\\
- Terminology or Proper Noun Issues (Terminology and Proper Noun Issues are related to inaccuracies when translating specialized vocabulary, inherent terms, and proper nouns from the source text.)\\
- Unnatural Style (Unnatural Style refers to translations that are grammatically correct but unnatural in the target language.)\\
- Ambiguity and Disambiguation (Ambiguity and Disambiguation errors occur when the ambiguities or errors in the source text, such as typographical errors, omissions, unclear abbreviations, and erroneous punctuation, are not faithfully reflected in the translation.)\\
- Buzzwords or Loanwords Issues (Buzzword or Loanword Issues occur when such terms are not translated accurately according to their usage in both the source and target languages. This includes the incorrect translation of popular sayings, newly created words, Internet slang, and memes.)\\
- other.\\
\\
Each error is classified as one of three categories: critical, major, and minor.\\

Critical errors inhibit comprehension of the text. Major errors disrupt the flow, but what the text is trying to say is still understandable. Minor errors are technically errors, but do not disrupt the flow or hinder comprehension.\\

The source text must be fully covered. Please only include errors and no spans that do not contain errors.\\

You will be given a full document and its translations, but only score one sentence at a time which is given in in <target\_segment></target\_segment> tags. \\

Overall quality score of the translation. After highlighting all errors, please choose the overall quality score.\\

The quality levels associated with numerical scores:\\

- 0: No meaning preserved: Nearly all information is lost in the translation.\\
- 33: Some meaning preserved: Some of the meaning is preserved but significant parts are missing. The narrative is hard to follow due to errors. The text may be phrased in an unnatural/awkward way. Grammar may be poor.\\
- 66: Most meaning preserved and few grammar mistakes: The translation retains most of the meaning. It may have some grammar mistakes or minor inconsistencies.\\
- 100: Perfect meaning and grammar: The meaning and grammar of the translation is completely consistent with the source. The text sounds like native text in the target language without any awkward phrases.\\

Use any number in the range between 0 and 100 for a fine-grained quality score.\\

Please score the following input \\
<input> \\
<source\_language>Korean</source\_language> \\
<source>\textbf{\textcolor{ForestGreen}{[KO]}}</source> \\
<target\_language>\textbf{\textcolor{ForestGreen}{[EN]}}</target\_language> \\
<translation>\textbf{\textcolor{ForestGreen}{[TRANSLATED RESULT]}}</translation> \\
<information>\textbf{\textcolor{ForestGreen}{[REF]}}</information>}\\
\end{tcolorbox}
\noindent\begin{minipage}{\linewidth}
\captionof{figure}{\textbf{Prompt used to evaluate six predefined error types with GPT‑5.1.} The template follows an FSP (Focus Sentence Prompting) format, providing task instructions, error definitions, and  score-range descriptors to guide the model’s judgments and output structure.}
\label{fig:eval_trans_prompt}
\end{minipage}
\end{center}

% \twocolumn
% \section{asd}

\end{document}